\documentclass[letterpaper]{article}            
\usepackage[draft]{aaai2026}               
\usepackage{times}                              
\usepackage{helvet}                             
\usepackage{courier}                            
\usepackage[hyphens]{url}                       
\usepackage{graphicx}                           
\usepackage{natbib}                             
\usepackage{caption}                            
\usepackage{pgfplotstable}
\usepackage{pgfplots}
\pgfplotsset{compat = newest}   

\usepackage{algorithm}
\usepackage{newfloat}
\usepackage{listings}
\DeclareCaptionStyle{ruled}{labelfont=normalfont,labelsep=colon,strut=off}
\floatstyle{ruled}
\newfloat{listing}{tb}{lst}
\floatname{listing}{Listing}

\usepackage[draft]{aaai2026}   

\usepackage{amsmath,amssymb,amsfonts,amsthm}

\usepackage{enumitem}

\usepackage{algorithm}
\usepackage{algpseudocode} 
\algnewcommand{\LeftComment}[1]{%
  \Statex\hspace*{-\algorithmicindent}\(\triangleright\) #1}

\usepackage{booktabs}    
\usepackage{nicefrac}    
\usepackage{microtype}   

\usepackage[utf8]{inputenc}
\usepackage[T1]{fontenc}

\usepackage{amsfonts}
\usepackage{footmisc}

\usepgfplotslibrary{groupplots}

\pgfplotsset{compat=1.17}
\usetikzlibrary{plotmarks}

\usepackage[utf8]{inputenc}
\usepackage{xcolor}
\usepackage{booktabs}
\usepackage{amsthm}
\usepackage{listings}
\usepackage{inconsolata}
\usepackage{tikz}
\usepackage{pgfplots}
\usepackage{amsmath}
\usepackage{microtype}
\usepackage{soul}

\usepgfplotslibrary{statistics,groupplots}
\usepackage{makecell}
\usepackage{float}

\usepackage{enumerate}%

\newcommand{\ac}[1]{\textcolor{red}{AC: #1}}

\newcommand{\popper}{\textsc{Popper}}

\theoremstyle{definition}
\newtheorem{definition}{Definition}

\lstnewenvironment{myalgorithm}[1][] 
{
    \lstset{ 
        showspaces=false,               
        showstringspaces=false,         
        mathescape=true,
        numbers=left,
        escapeinside={*}{*},
        columns=flexible,
        numbersep=2pt,        
        keywordstyle=\bfseries,
        keywords={,and, return, not, def, in, if, else, for, foreach, while, }
        numbers=left,
        xleftmargin=.0\textwidth,
        #1 
    }
}
{}

\title{Symbolic Snapshot Ensembles}

\author{
Mingyue Liu$^{1}$, 
Andrew Cropper$^{1}$\\
}

\affiliations{
$^{1}$University of Oxford, Department of Computer Science\\
\texttt{rita.liu@cs.ox.ac.uk}, \texttt{andrew.cropper@cs.ox.ac.uk}
}

\begin{document}
\maketitle

\begin{abstract}
  Inductive logic programming (ILP) is a form of logical machine learning.
  Most ILP algorithms learn a single hypothesis from a single training run.
  Ensemble methods train an ILP algorithm multiple times to learn multiple hypotheses.
  In this paper, we train an ILP algorithm only once and save intermediate hypotheses.
  We then combine the hypotheses using a minimum description length weighting scheme.
  Our experiments on multiple benchmarks, including game playing and visual reasoning, show that our approach improves predictive accuracy by 4\% with less than 1\% computational overhead.


\end{abstract}
\section{Introduction}

Inductive logic programming (ILP) is a form of machine learning \cite{mugg:ilp,ilpintro}. 
The goal is to search for a hypothesis that generalises given training examples and background knowledge (BK).
The distinguishing feature of ILP is that it uses logical rules to represent hypotheses, examples, and BK.

As with other forms of machine learning, ILP ensembles can reduce generalisation error \cite{quinlan1996boosting,kramer2001demand,hoche2001relational,jiang2006boosting,salvini2007skimmed}.
For instance, \citet{ilpbagging2003} use bagging to learn multiple hypotheses from subsets of the training data and combine them via majority voting. 
Their empirical results show that their method improves accuracy by 3-4\% compared to a single hypothesis prediction.

Existing ensemble approaches have two notable limitations.
First, when learning from a small dataset, bootstrap samples can significantly overlap, reducing ensemble diversity and limiting effectiveness ~\citep{dietterich2000ensemble}. 
Second, training an ILP algorithm multiple times to build an ensemble is computationally expensive.

In this paper, rather than train an ILP algorithm multiple times, we train an algorithm only once and save the intermediate hypotheses it finds.
When the ILP algorithm finds a hypothesis whose cost is lower than any encountered so far, we add it to a snapshot pool and we ask the ILP algorithm to keep searching until it exceeds a given timeout.
We then combine the hypotheses using a minimum description length (MDL) \cite{mdl} weighting scheme. 
We use the weighted combination to make predictions.
We follow \citet{huang2017snapshot} and call this approach a \emph{snapshot} ensemble.

\subsubsection*{Novelty and Contributions}

The novelty of this paper is using multiple hypotheses learned by a \textit{single} ILP run to build a snapshot ensemble and combining the hypotheses with an MDL weighting scheme.  
The impact, which we demonstrate on multiple datasets, including game-playing and visual reasoning, is a reduction in generalisation error with almost no extra training cost compared to the classical bagging approach.
Overall, we contribute the following:

\begin{itemize}
    \item 
          We describe the first single-run ensemble framework for ILP, which is sufficiently general to work with any anytime ILP algorithm.
    \item 
          We introduce an MDL-based weighting scheme that balances training fit against ensemble complexity.
    \item 
    We experimentally show on multiple domains, including visual reasoning and game playing, that our snapshot ensemble approach improves predictive accuracy by 4\% with less than 1\% overhead.
    \item We experimentally show that our ensemble approach matches or surpasses the predictive performance of standard bagging methods, while requiring significantly less computational cost.
\end{itemize}

\section{Related Work}

\paragraph{ILP.}
Most ILP approaches learn a single hypothesis from one run on the training data \cite{foil,progol,aleph,metagold,tilde,ilasp,dilp,popper}.
For instance, ProbFOIL~\cite{probfoil} extends FOIL \cite{foil} to 
find the single best probabilistic hypothesis.
We differ from most ILP approaches by saving hypotheses found during training and combining them into an ensemble.
\paragraph{ILP ensembles.}
Ensembling in ILP focuses on bagging and boosting \cite{quinlan1996boosting,kramer2001demand,hoche2001relational,ilpbagging2003,jiang2006boosting,salvini2007skimmed}.
These approaches require multiple training runs or resampling procedures.
For instance, \citet{ilpbagging2003} train an ILP learner on multiple bootstrap samples of the data (sampling with replacement) and combine the induced hypotheses by majority vote to improve generalisation. 
To reduce computational overhead of bagging, \citet{salvini2007skimmed} ensemble single rules rather than full hypotheses (with multiple rules), trading a slight decrease in accuracy for substantially faster training. 
\citet{yan2024explainable} introduce a method that merges relational decision trees \cite{tilde} ensemble candidates into a single compressed decision list, reducing the hypothesis's complexity while retaining the predictive accuracy of bagging or boosting.
In contrast to these ensemble methods, our approach requires only a single ILP run.
As far as we know, no previous work describes a single-run ensemble method in ILP.

\paragraph{Single-run ensembles.}
Outside ILP, some ensemble methods capture multiple hypotheses from a single training run. 
Horizontal voting ensembles \cite{xie2013horizontal} and checkpoint ensembles \cite{chen2017checkpoint} both store intermediate checkpoints at various training epochs and combine their predictions, while snapshot ensembles use a cyclic learning-rate schedule that forces the model into different local optima \cite{huang2017snapshot}.
Another direction of research explores the construction of ensembles through subnetwork-based methods. For instance, MotherNet \cite{wasay2020mothernets} expands a trained parent network by adding layers and neurons to form child networks. FreeTickets \cite{liu2021deep} preserves diverse and low-cost sparse subnetworks during training. Prune and Tune Ensemble \cite{whitaker2022prune} creates ensemble diversity by pruning each cloned network independently.
Our method differs from these network‐based approaches because it does not rely on manipulating a continuous weight space (e.g.\ via cyclic learning rates or pruning).
Instead, we harness the natural diversity of structurally distinct rule sets generated during ILP’s search.

\section{Problem Setting}
We now define our problem setting.
We assume familiarity with logic programming \cite{lloyd:book,luc:book}.
\subsection{Inductive Logic Programming}
We define an ILP input:
\begin{definition}[\textbf{ILP input}]
\label{def:probin}
An ILP input is a tuple $(E, B, \mathcal{H})$ where $E=(E^+,E^-)$ is a pair of sets of ground atoms representing positive ($E^+$) and negative ($E^-$) examples, $B$ is a definite program representing background knowledge, and $\mathcal{H}$ is a hypothesis space, i.e. a set of possible hypotheses.
\end{definition}
\noindent
We restrict hypotheses and background knowledge to definite programs with the least Herbrand model semantics.

We define a cost function:
\begin{definition}[\textbf{Cost function}]
\label{def:cost_function}
Given an ILP input $(E, B, \mathcal{H})$, a cost function $cost_{E,B}~:~\mathcal{H}~\mapsto~\mathbb{N}$ assigns a numerical cost to each hypothesis $h \in \mathcal{H}$.
\end{definition}

\noindent
The MDL principle \cite{mdl} trades off hypothesis complexity (hypothesis size) and data fit (training accuracy). 
The MDL principle can be expressed as finding a hypothesis that minimises $L(h)+L(E|h)$, where $L(h)$ is the syntactic complexity of the hypothesis $h$ and $L(E|h)$ is the complexity of the data when encoded using $h$.
Following \citet{maxsynth}, we evaluate $L(E|h)$ as the cost of encoding the exceptions to the hypothesis, i.e. the number of false positives and false negatives. 
Given a hypothesis $h$, background knowledge $B$, and a set of examples $E$, 
a \emph{false positive} is a negative example in $E$ entailed by $h \cup B$ and a \emph{false negative} is a positive example in $E$ not entailed by $h \cup B$.
We denote the number of false positives and false negatives as $fp_{E}(h)$ and $fn_{E}(h)$  respectively. 
We evaluate the complexity of the hypothesis $complexity(h)$ as its size. 
Specifically, we define the \emph{MDL} cost function as:
\begin{align*}
MDL(h) = fp_{E}(h)+fn_{E}(h)+complexity(h)
\end{align*}



\noindent
 The standard ILP problem is: 
\begin{definition}[\textbf{ILP problem}]
\label{def:opthyp}
Given an ILP input $(E, B, \mathcal{H})$ and a cost function \emph{cost$_{E,B}$}, 
the ILP problem is to find an \emph{training-optimal hypothesis} $h \in \mathcal{H}$ that minimises the cost function.
\end{definition}





\subsection{Ensemble Inductive Logic Programming}
\noindent
 The standard ILP problem finds a single hypothesis.
 We extend the ILP setting to find an ensemble of hypotheses:
\begin{definition}[\textbf{Ensemble ILP input}]
\label{def:probin}
An ensemble ILP input is a tuple $(E, B, \mathcal{H}, \mathcal{M})$ where $E$, $B$, and $\mathcal{H}$ are as in a standard ILP input and 
$\mathcal{M}$ is an \emph{ensemble method} that combines hypotheses to make a prediction.

\end{definition}

\noindent 
We define an ensemble cost function:
\begin{definition}[\textbf{Ensemble cost function}]
\label{def:cost_function}
Given 
an ILP input $(E, B, \mathcal{H})$, 
a cost function $cost_{E,B,\mathcal{M}} : \mathcal{P(H)} \mapsto \mathbb{N}$ 
assigns a numerical cost to each ensemble.
\end{definition}

\noindent
We define \emph{ensemble ILP}:

\begin{definition}[\textbf{Ensemble ILP problem}]
Given an ensemble ILP input $(E, B, \mathcal{H}, \mathcal{M})$ and an ensemble cost function \emph{cost$_{E,B,\mathcal{M}}$}, the \emph{ensemble ILP} problem is to find an optimal ensemble that minimises the ensemble cost function.
\end{definition}

\noindent
Ensemble ILP extends classical ILP by finding a set of hypotheses whose combined predictions via an ensemble method minimises a given cost function on the training data.

\section{Snapshot Ensembles}
\label{sec:algo}
We now describe our snapshot ensemble method.
Our method has three phases: building a snapshot pool of hypotheses, building a weighted combination of hypotheses, and using the weighted combination to make predictions.
We describe each phrase in turn.

\subsection*{Phase 1: Snapshot Pool}

The goal of Phase~1 is to build a snapshot pool of hypotheses for an ensemble.
Algorithm~\ref{alg:snapshot-pool-collection} shows the procedure.
This function takes as input an ILP algorithm $\mathcal{I}$, training examples $E$, background knowledge $B$, and a cost function \emph{cost} that measures the quality of each hypothesis (Definition~\ref{def:cost_function}).
We require that \(\mathcal{I}\) is an anytime algorithm that returns multiple candidate hypotheses whilst it searches for an optimal hypothesis. 
The \(\mathrm{cost}\) parameter can represent various scoring measures, such as those listed in Section \ref{cost_functions}.



Algorithm~\ref{alg:snapshot-pool-collection} begins by asking \(\mathcal{I}\) for the next candidate hypothesis in line 4.
Whenever $\mathcal{I}$ returns a hypothesis $h$ whose cost is \emph{no larger} than the best cost seen so far, $h$ is added to the snapshot pool~$\mathcal{S}$.
The loop now runs until the timeout~$\tau$ expires, ensuring that hypotheses discovered within the time limit are saved.
The output is a snapshot pool of candidate hypotheses, each of which was the best at some point during the search.

\begin{algorithm}[ht!]
\caption{\textsc{SnapshotPoolCollection} (Phase 1)
}
\label{alg:snapshot-pool-collection}
\begin{algorithmic}[1]
\Require ILP system $\mathcal{I}$, cost function $\mathrm{cost}$, background knowledge $B$, training examples $E$, timeout $\tau$
\Ensure A pool of candidate hypotheses \(\text{$\mathcal{S}$}\)
\Function{SnapshotPoolCollection}{$\mathcal{I}$, $\mathrm{cost}$, $B$, $E$, $\tau$}
    \State $\text{$\mathcal{S}$} \gets \emptyset$
    \State $\text{bestCost} \gets +\infty$
    \State $t_{0} \gets \textsc{TimeNow}()$
    \While{$\textsc{TimeNow}() - t_{0} < \tau$}
        \State $h \gets \textsc{NextCandidate}(\mathcal{I})$
        \State $c \gets \mathrm{cost}(h,B,E)$
        \If{$c \le \text{bestCost}$}      
            \State $\text{$\mathcal{S}$} \gets \text{$\mathcal{S}$} \cup \{h\}$
            \If{$c < \text{bestCost}$}   
                \State $\text{bestCost} \gets c$
            \EndIf
        \EndIf
    \EndWhile
    \State \Return $\text{$\mathcal{S}$}$
\EndFunction
\end{algorithmic}
\end{algorithm}

\subsection*{Phase 2: Weighted Combination}

The goal of Phase 2 is to assign a normalised weight \(w_h\) to every hypothesis \(h\in\mathcal{S}\), thus converting the snapshot pool into a weighted ensemble.
Algorithm~\ref{alg:weight-assignment-bayes} describes this process.
It takes as input the snapshot pool \(\mathcal{S}\), background knowledge \(B\), training examples \(E\), and hyper-parameters \(\alpha,\beta>0\).
It returns an array of normalised weights $w$ for the hypotheses in the snapshot pool.
We treat each hypothesis \(h_i \in \mathcal{S}\) as having a prior penalising hypothesis complexity and a likelihood derived from its empirical coverage on the training set. Specifically, we define the fraction of correctly classified training examples as:
\[
  \mathrm{coverage}(h_i) 
  \;=\; \frac{\mathrm{tp}(h_i) + \mathrm{tn}(h_i)}{\mathrm{tp}(h_i) + \mathrm{tn}(h_i) + \mathrm{fp}(h_i) + \mathrm{fn}(h_i)}
\]
We also define the unnormalised posterior probability of \(h_i\) as:
\[
raw_i
=\underbrace{\bigl(\mathrm{coverage}(h_i)\bigr)^{\beta}}_{\text{likelihood}}
 \;\times\;
 \underbrace{\exp\bigl(-\alpha\,\mathrm{mdl}(h_i)\bigr)}_{\text{prior}}
\]
where \(\alpha \ge 0\) and \(\beta \ge 0\) are hyperparameters to balance hypothesis coverage against MDL cost, and $raw$ is the value of weights before normalisation. 
We use MDL in the prior because it penalises both absolute error $(\mathrm{fp}+\mathrm{fn})$ and the complexity $\mathrm{complexity}(h_i)$.  
A size-only penalty would overlook high-error but short hypotheses (underfitting) until the likelihood acts, while a pure MDL weight solely would conflate absolute and proportional error.
By separating the roles, $\alpha$ controls an a priori cost on total error + complexity and $\beta$ rewards the fraction of correctly classified examples.
We then normalise these weights:
\[
    w_i \;=\; \frac{raw_i}{\sum_{j=1}^{k} raw_j}
    \quad\text{so}\quad \sum_{i=1}^{k} w_i \;=\; 1,
\]
where $k$ is the size of the snapshot pool.
This normalisation forms a probability distribution over the snapshot pool \(\mathcal{S}\), with higher likelihood given to hypotheses exhibiting strong coverage and lower complexity.


\begin{algorithm}[ht!]
\caption{\textsc{WeightAssignmentAndNormalisation} (Phase 2)}
\label{alg:weight-assignment-bayes}
\begin{algorithmic}[1]
\Require Snapshot pool $\mathcal{S}$, background knowledge $B$, training examples $E$, hyperparameters $\alpha,\beta >0$
\Ensure An array of weights $w$
\Function{AssignWeights}{$\mathcal{S}$, $B, E$, $\alpha$, $\beta$}
    \State $\text{w}\gets \text{array of length } |\mathcal{S}| $

    \State $\text{sumWeights}\gets 0$
    \ForAll{$h\in\mathcal{S}$}
        \State $\mathrm{cov}\gets\textsc{Coverage}(h, B, E)$
        \State $\mathrm{mdl}\gets\textsc{MDLScore}(h, B, E)$
        \State $\text{w}[h]\gets(\mathrm{cov})^{\beta}\,\exp(-\alpha\,\mathrm{mdl})$
        \State $\text{sumWeights}\gets\text{sumWeights}+\text{w}[h]$
    \EndFor
    \ForAll{$h\in\mathcal{S}$}
        \State $\text{w}[h]\gets \text{w}[h] / \text{sumWeights}$
    \EndFor
    \State\Return $\text{w}$
\EndFunction
\end{algorithmic}
\end{algorithm}

\subsection*{Phase 3: Ensemble Prediction}
Given a snapshot pool $\mathcal{S}$ from Phase~1, normalised weights $w$ from Phase~2, and an unseen test example $x$, the goal of Phase 2 is to predict the binary classification $y(x) \in \{1,0\}$ of $x$.
Each hypothesis $h \in \mathcal{S}$ returns a binary prediction \(y_h(x)\in\{0,1\}\). 
The ensemble's prediction for \(x\) is computed as:
\[
    y(x) = 
    \begin{cases}
    1 & \text{if} \quad (\sum_{h\in\mathcal{S}}w_h\,y_h(x))\ge0.5 \\
    0 & \text{otherwise.}
    \end{cases}
\]

\noindent
Thus, hypotheses with higher assigned weights have greater influence on the final ensemble prediction, while all hypotheses contribute according to their weights.

\section{Experiments}

To test our claim that our snapshot ensemble can reduce generalisation error, our experiments aim to answer the question:


\begin{description}
\item[Q1] Can snapshot ensembles improve generalisation?
\end{description}

\noindent
To answer \textbf{Q1} we compare the predictive accuracy of an anytime ILP algorithm with and without our snapshot ensemble idea.
In other words, we compare the predictive accuracy of the final training-optimal hypothesis (Definition \ref{def:opthyp}) found by an ILP system against our snapshot ensemble method.


Our ensemble method allows a snapshot pool to include multiple hypotheses with the same cost, notably multiple training-optimal hypotheses.
To understand the impact of including multiple hypotheses, our experiments aim to answer the question:
\begin{description}
\item[Q2] Can finding multiple optimal hypotheses improve snapshot ensemble generalisation?
\end{description}

\noindent
To answer \textbf{Q2} we compare the predictive accuracy of our snapshot ensemble idea with and without multiple optimal hypotheses.

Our ensemble method adds non-optimal hypotheses to a snapshot pool. 
To understand the impact of including non-optimal hypotheses, our experiments aim to answer the question:

\begin{description}
\item[Q3] Can including non-optimal hypotheses improve snapshot ensemble generalisation?
\end{description}

\noindent
To answer \textbf{Q3} we compare the predictive accuracy of snapshot ensemble idea with and without non-optimal hypotheses.



In addition to accuracy, we claim that a key advantage of snapshot ensembles is that they require little overhead, as we train an ILP algorithm only once.
To evaluate this claim, our experiments aim to answer the question:

\begin{description}
\item[Q4] What is the overhead of snapshot ensembles?
\end{description}

\noindent
To answer \textbf{Q4}, we measure the overhead of our approach compared to a standard anytime ILP algorithm.
Since our approach requires only a single ILP run and incurs no additional cost during snapshot pool collection (Phase 1), we focus on the overhead from adding weights (Phase 2) and ensemble prediction (Phase 3). 
We report this overhead using the largest possible snapshot pool, considering both non-optimal hypotheses and scenarios with multiple optimal hypotheses.

Finally, to position our snapshot ensemble method within the broader symbolic ensemble literature, we aim to answer the question:

\begin{description}
    \item[Q5] How does our snapshot ensemble method compare to existing bagging methods?
\end{description}

\noindent
To answer \textbf{Q5}, we compare the predictive accuracy and computational overhead of our snapshot ensemble against a bagging method.

\paragraph{Reproducibility.}
The code and experimental data for reproducing the experiments are available as supplementary material and will be made publicly available if the paper is accepted for publication.    

\subsection{Methods}

\subsubsection{Evaluation Metric}
For each task, we record test-set accuracies.
\(ACC_{\text{snap}}\) is the  snapshot‐ensemble accuracy.
\(ACC_{\text{base}}\) is the accuracy of the single training‐optimal hypothesis and is the baseline accuracy.

\subsubsection{Significance}
We repeat each learning task 6 times on a single CPU of an 8-core 3.2 GHz Apple M1 machine.
We report the mean performance along with 95\% confidence intervals.

\subsubsection{ILP System}
Although our snapshot ensemble method is general and system agnostic, we demonstrate it with the ILP system \popper{} \cite{popper} V4.4.0.
We use this system because it is an anytime approach and learns from noisy data \cite{maxsynth}.
\popper{} iteratively learns hypotheses that minimise a cost function.
Each time \popper{} finds a better hypothesis, it outputs the hypothesis.
Given sufficient time, \popper{} is guaranteed to learn an optimal hypothesis.
However, in practice, it takes a long time to find an optimal hypothesis.
Therefore, we use a 3600 seconds timeout and use the final hypothesis found by \popper{}.
We refer to this final hypothesis as the \textsc{baseline}.

\subsubsection{Bagging Ensemble}
To answer \textbf{Q5}, we implement bagging using \popper{} under the same conditions as our snapshot method. 
We use the method of \citet{ilpbagging2003} and \citet{salvini2007skimmed}.
We ensemble entire hypotheses rather than individual rules, since \citet{salvini2007skimmed} report that this achieved the best predictive accuracy. 
We construct each bag by sampling $K$ examples uniformly at random with replacement from the original training set of size $K$, so that individual examples may be repeated across bags.
For full reproducibility, we fix the random seed for each bag's replacement at 43,44,45.
For each learning task, we generate 3 bootstrap samples of training examples while keeping the background knowledge unchanged. Each dataset is used to train \popper{} independently for 3600 seconds, yielding 3 hypotheses. 
We assign uniform weights to hypotheses.



\subsubsection{Cost Functions}
\label{cost_functions}
 We consider three cost functions in Phase 1.
We write $\mathrm{Lex}(x_1,\dots,x_k)$ for \emph{lexicographic ordering}: two tuples are compared by their first component, then, only if those are equal, by the second argument, and so on.

\begin{itemize}
    \item \textbf{ErrorSize}: minimises a lexicographic combination of error and hypothesis complexity:
    \[
    \mathrm{errorsize}(h) \;=\; \mathrm{Lex}\!\bigl(\,\mathrm{fp}(h) + \mathrm{fn}(h),\; \mathrm{complexity}(h)\bigr).
    \]
    \item \textbf{MDL}: as defined in Definition~\ref{def:cost_function}.
    \item \textbf{Lexfnsize}: minimises a lexicographic ordering of false negatives, false positives and hypothesis complexity:
    \[
    \mathrm{Lexfnsize}(h) \;=\; \mathrm{Lex}(\mathrm{fn}(h),\; \mathrm{fp}(h),\; \mathrm{complexity}(h)).
    \]

\end{itemize}

\subsubsection{Datasets}
We use the following datasets.
For each dataset, we split the examples into training, validation, and testing sets with a ratio of $7:2:1$.

 \textbf{1D-ARC.} The \emph{1D-ARC} dataset \cite{onedarc} is a one-dimensional adaptation of the Abstraction and Reasoning Corpus (ARC) \cite{arc}.

\textbf{Alzheimer.}
These real-world tasks \cite{alzheimer} involve learning rules describing four properties desirable for drug design against Alzheimer's disease.

\textbf{Noisy-dropk.}
Noisy-dropk is a list transformation task adding 20\% noisy data \cite{maxsynth}.



\textbf{Wn18rr.} Wn18rr \cite{dettmers2018convolutional} is a widely used real-world knowledge base with 11 relations from WordNet.

\textbf{Yeast.} The goal of this task is to learn rules describing whether a yeast gene codes for a protein involved in the general function category of metabolism \cite{davis2005integrated}. 


\textbf{IGGP.} In inductive general game playing (IGGP) \cite{iggp}, the task is to induce rules from game traces.

\textbf{Drug-Drug.} Given each drug’s enzymes, transporters, and targets, the task is to predict if a drug pair interacts \cite{dhami2018drug}.

There are 37 tasks.
We consider three cost functions for each task.
Therefore, there are 111 tasks in total.





\subsubsection{Hyperparameters}
We tune the two ensemble hyperparameters \(\alpha\) and \(\beta\) to balance hypothesis coverage against MDL cost. Recall from Section~\ref{sec:algo} that each snapshot \(h_i\) receives weight:
\[
raw_i
={\bigl(\mathrm{coverage}(h_i)\bigr)^{\beta}}
 \;\times\;
 {\exp\bigl(-\alpha\,\mathrm{mdl}(h_i)\bigr)}
\]
\noindent
Taking the natural logarithm gives:
\[
\ln raw_i
\;=\;
\beta\,\ln\bigl(\mathrm{coverage}(h_i)\bigr)
\;-\;\alpha\,\mathrm{mdl}(h_i)
\]

\noindent
Therefore, coverage contributes \(\beta\,\ln(\mathrm{coverage})\) and cost contributes \(-\alpha\,\mathrm{mdl}\) to the log‑weight.

To quantify the trade‐off between coverage and cost, we define the \emph{coverage–cost ratio}:
\[
  R \;=\; \frac{\beta\,\ln\bigl(\mathrm{coverage}_{\max}/\mathrm{coverage}_{\min}\bigr)}{\alpha\,\bigl(\mathrm{mdl}_{\max}-\mathrm{mdl}_{\min}\bigr)}
\]
where \(\mathrm{coverage}_{\max}\) and \(\mathrm{coverage}_{\min}\) are the approximate maximum and minimum coverages, and \(\mathrm{mdl}_{\max}\) and \(\mathrm{mdl}_{\min}\) are the approximate maximum and minimum MDL costs across all snapshots observed during training.
Intuitively, coverage and cost contribute equally when $R = 1$.
A bigger $R$ indicates the weights are more coverage-dominant.
We therefore fix \(\beta=2\) and sweep 
\(\alpha\in\{0.0005,0.001,0.0013,0.0017,0.005,0.03,0.06\}\),
which yields \(\;R\in\{10,5,4,3,1,0.2,0.08\}\).
These values of \(R\) range from strong coverage dominance through balance to strong MDL‐cost dominance. 
The values \(\alpha=0.0017\) (\(R=3\)) provide the best validation‐set accuracy averaged among all cost functions and all tasks for 1 trial.
Under this setting, the MDL term accounts for about 25 \% of the total variation in $\ln raw_i$.
Accordingly, all reported results in Section \ref{'result'} use \(\beta=2\) and \(\alpha=0.0017\).



\begin{figure}[!b]
\centering
\begin{tikzpicture}

\begin{groupplot}[
  group style={
      group size=1 by 3,
      vertical sep=1cm,
      x descriptions at=edge bottom,
      y descriptions at=edge left
  },
  width=\columnwidth,
  height=.50\columnwidth,
  xmin=-1, xmax=13,     
  ymin=-5, ymax=10,
  xtick=\empty,
  xlabel={Task},
  every axis plot/.append style={mark size=0.8pt},
]

\nextgroupplot[
      title={Cost Function: \textsc{MDL}},
      ylabel={Accuracy difference (\%)},
            xmin=-1, xmax=13,   
      ymin=-5, ymax=8
]
    \addplot [only marks, mark=*, black, fill opacity=0.6,
            error bars/.cd, y dir=both, y explicit,
            error bar style={color=red}]
      table [x=n, y=mean, y error minus=minus, y error plus=plus]
      {diff_ranked_mdl.txt};
  \addplot [dashed, domain=0:24, samples=2, color=black] {0};

\nextgroupplot[
      title={Cost Function: \textsc{Errorsize}},
      ylabel={}
]
  \addplot [only marks, mark=*, black, fill opacity=0.6,
            error bars/.cd, y dir=both, y explicit,
            error bar style={color=red}]
      table [x=n, y=mean, y error minus=minus, y error plus=plus]
      {diff_ranked_errorsize.txt};
\addplot [dashed, domain=0:24, samples=2, color=black] {0};

\nextgroupplot[
      title={Cost Function: \textsc{Lexfnsize}},
      ylabel={},
      xticklabel style={font=\scriptsize},
      xmin=-1, xmax=13,   
      ymin=-10, ymax=49
]
  \addplot [only marks, mark=*, black, fill opacity=0.6,
            error bars/.cd, y dir=both, y explicit,
            error bar style={color=red}]
      table [x=n, y=mean, y error minus=minus, y error plus=plus]
      {diff_ranked_lexfnsize.txt};
\addplot [dashed, domain=0:24, samples=2, color=black] {0};

\end{groupplot}
\end{tikzpicture}
\caption{Task-level change in accuracy (our symbolic ensemble's predictive accuracy minus the baseline's) across datasets. 
}
\label{fig:q1_single_column_three}
\end{figure}
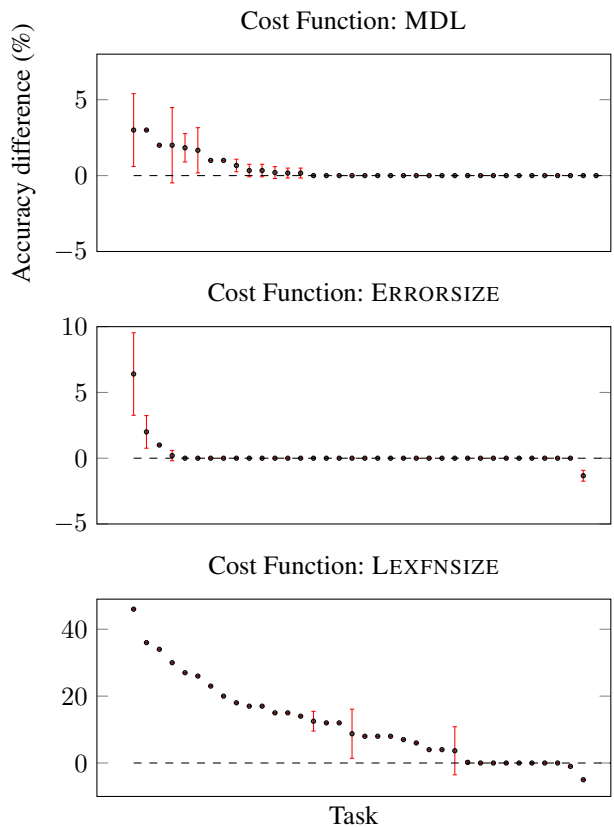

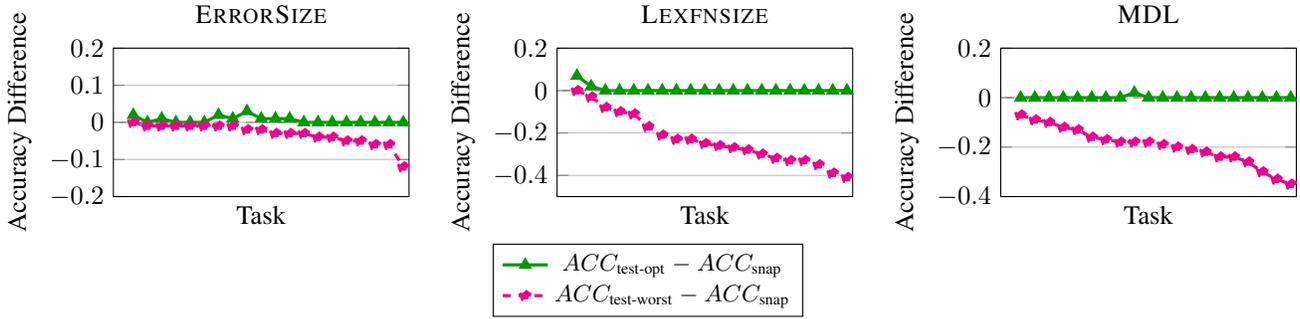
\begin{figure*}[ht!]
  \centering
  \pgfplotstableread{diff_testopt_minus_ensemble_errorsize_sort.txt}\dataTestErr
  \pgfplotstableread{diff_worst_minus_ensemble_errorsize_sort.txt}\dataWorstErr
  \pgfplotstableread{diff_testopt_minus_ensemble_fnsize_sort.txt}\dataTestFn
  \pgfplotstableread{diff_worst_minus_ensemble_fnsize_sort.txt}\dataWorstFn
  \pgfplotstableread{diff_testopt_minus_ensemble_mdl_sort.txt}\dataTestMdl
  \pgfplotstableread{diff_worst_minus_ensemble_mdl_sort.txt}\dataWorstMdl

  \begin{tikzpicture}
    \begin{groupplot}[
        group style={
          group size=3 by 1,
          horizontal sep=5.6em,
        },
        width=0.31\textwidth,
        height=0.2\textwidth,
        xlabel={Task},
        ylabel={ Accuracy Difference},
        xtick=\empty,
        enlarge x limits=0.02,
        ymajorgrids
      ]

      \nextgroupplot[
        title={\textsc{ErrorSize}},
        ymin=-0.2, ymax=0.2,
        legend to name=sharedlegend,
        legend style={
          at={(0.5,-0.18)},
          anchor=north,
          legend columns=1,
          font=\footnotesize
        }
      ]
      \addplot[very thick, green!60!black, mark=triangle*, mark options={scale=0.9}]
        table[x expr=\coordindex+1, y index=0]{\dataTestErr};
      \addlegendentry{$ACC_\text{test‐opt}$ $-$ $ ACC_\text{snap}$}
      \addplot[very thick, magenta, dashed, mark=pentagon*, mark options={scale=0.9}]
        table[x expr=\coordindex+1, y index=0]{\dataWorstErr};
      \addlegendentry{$ACC_\text{test-worst}$ $-$ $ ACC_\text{snap}$}
      \addplot[dashed, gray] coordinates {(0,0) (20,0)};

      \nextgroupplot[
        title={\textsc{Lexfnsize}},
        ymin=-0.50, ymax=0.2
      ]
      \addplot[very thick, green!60!black, mark=triangle*, mark options={scale=0.9}]
        table[x expr=\coordindex+1, y index=0]{\dataTestFn};
      \addplot[very thick, magenta, dashed, mark=pentagon*, mark options={scale=0.9}]
        table[x expr=\coordindex+1, y index=0]{\dataWorstFn};
      \addplot[dashed, gray] coordinates {(0,0) (20,0)};

      \nextgroupplot[
        title={\textsc{MDL}},
        ymin=-0.4, ymax=0.2
      ]
      \addplot[very thick, green!60!black, mark=triangle*, mark options={scale=0.9}]
        table[x expr=\coordindex+1, y index=0]{\dataTestMdl};
      \addplot[very thick, magenta, dashed, mark=pentagon*, mark options={scale=0.9}]
        table[x expr=\coordindex+1, y index=0]{\dataWorstMdl};
      \addplot[dashed, gray] coordinates {(0,0) (20,0)};

    \end{groupplot}
  \end{tikzpicture}

  \pgfplotslegendfromname{sharedlegend}

\caption{
Per-task differences in accuracy between (1) the best individual snapshot (test-optimal) and the snapshot ensemble, and (2) the worst individual hypothesis and the snapshot ensemble, under three cost functions.}
  \label{fig:delta-ensemble-subplots}
\end{figure*}

\subsection{Results}
\label{'result'}
\subsubsection{Q1. Can Snapshot Ensembles Improve Generalisation?}

Figure \ref{fig:q1_single_column_three} shows the differences in predictive accuracies when using snapshot ensembles compared to the baseline.
Our snapshot ensemble matches or improves on the baseline on nearly every task.  
A t-test or a Wilcoxon Signed-Rank Test (depending on whether the data is normal) confirms $(p < 0.05)$ that our approach improves accuracy on 45/111 (41\%) tasks and reduces accuracy on 3/111 (3\%) tasks.
For \textsc{ErrorSize} the mean improvement is $0.2\%\pm0.4\%$ and the ensemble underperforms on only one task.   
For \textsc{mdl} the mean improvement is $0.5\%\pm0.3\%$ without a negative case.
The largest improvement is for \textsc{LexFnSize} where accuracy improves by $11.8\%\pm4.0\%$ and the ensemble wins on 27/37 ($73\%$) tasks.

These results show that our ensemble approach leads to substantial improvements with the \textsc{Lexfnsize} cost function but more modest improvements with the \textsc{mdl} and \textsc{Errorsize} functions.
The reason for the difference is that the baseline ILP algorithm with the \textsc{mdl} and \textsc{Errorsize} cost functions already performs well and rarely overfits.

To illustrate this lack of overfitting, we define \(ACC_{\text{test--opt}}\) as the best test accuracy achieved by \emph{any} intermediate hypothesis found by \popper{}.
We measure the \emph{over-fit gap} as \(ACC_{\text{test--opt}} - ACC_{\text{base}}\), where a positive value indicates that the final training-optimal hypothesis overfits.
For \textsc{Errorsize}, the over-fit gap is negligible (\textbf{$0.0\% \pm 0.0\%$}).
It is also small for \textsc{mdl} (\textbf{$0.5\% \pm 0.9\%$}).
These results suggest that these cost functions help \popper{} avoid severe overfitting. 
By contrast, for \textsc{Lexfnsize}, the gap is consistently large (\textbf{$12\% \pm 10\%$}), indicating that the training-optimal hypotheses often generalise poorly.

\begin{figure}[h!]
  \centering
  \begin{tikzpicture}
    \begin{axis}[
      width=0.8\columnwidth,
      height=0.4\columnwidth,
      scale only axis,
      xlabel={Overfit Gap (\%)},
      ylabel={Snapshot Improvement (\%)},
      xlabel near ticks,
      ylabel near ticks,
      tick label style={font=\footnotesize},
      label style={font=\footnotesize},
      axis line style={thick},
      grid=major,
      grid style={dashed, gray!30},
      legend style={
        font=\footnotesize,
        draw=none,
        at={(0.02,0.98)},
        anchor=north west
      },
    ]
      \addplot[
        only marks,
        mark=*,
        mark size=1.5pt,
        draw=black,
        fill=black!50,
      ] table [x=overfit, y=improvement] {
overfit improvement
0.39 2.00
0.68 0.00
-0.14 0.00
0.00 0.00
-0.32 0.00
3.28 3.00
0.00 0.00
0.00 0.00
0.00 0.00
0.38 6.40
7.01 0.00
0.00 0.00
0.47 1.00
0.06 0.00
0.88 1.00
0.00 0.00
1.41 0.20
0.00 0.00
-0.32 0.00
12.62 8.75
0.23 0.00
0.14 0.00
10.66 3.67
1.23 0.17
0.17 0.00
14.45 12.50
2.74 1.67
0.24 0.00
9.48 8.00
-0.02 0.00
-0.27 0.00
11.61 12.00
1.88 2.00
-0.30 0.00
14.55 4.00
1.33 2.00
-0.14 0.00
20.00 20.00
0.00 0.00
0.18 0.00
6.36 6.00
1.33 1.83
0.11 0.00
17.27 17.00
2.27 0.00
-0.36 0.00
13.64 14.00
-0.09 0.00
0.38 0.00
3.85 -1.00
0.46 0.00
-0.38 0.00
1.92 -5.00
0.27 0.00
-0.26 0.00
7.69 8.00
0.23 0.00
-0.35 0.00
7.69 4.00
1.27 1.00
0.35 0.00
7.69 8.00
0.42 0.33
0.38 0.00
15.49 15.00
0.09 0.17
0.31 0.00
23.24 23.00
-0.27 0.00
-0.11 0.00
16.90 17.00
0.46 0.00
0.01 0.00
11.97 12.00
-0.15 0.00
-0.06 0.00
18.31 18.00
0.27 0.67
-0.21 0.00
-0.08 0.00
-0.21 0.00
0.00 0.00
26.50 27.00
0.00 0.00
0.01 0.20
34.38 34.00
0.03 0.00
0.00 0.00
35.89 36.00
0.00 0.00
-0.01 0.00
0.12 0.00
-0.02 0.00
0.00 0.00
30.43 30.00
0.00 0.00
-0.26 0.00
8.38 7.00
-0.39 3.00
0.05 0.00
-0.04 0.00
0.33 -0.17
26.44 26.00
-0.09 -1.33
15.48 15.00
0.73 0.33
-0.02 -0.17
46.08 46.00
-0.07 0.00
-0.29 0.00
-0.29 0.00
      };


\addplot[
  no markers,
  very thick,
  color=red,
  domain=-1:50,      
  samples=2,         
] {0.9663*x - 0.253};

    \end{axis}
  \end{tikzpicture}
  \caption{Snapshot improvement vs.\ overfit gap with linear regression line.}
  \label{fig:overfit-scatter}
\end{figure}
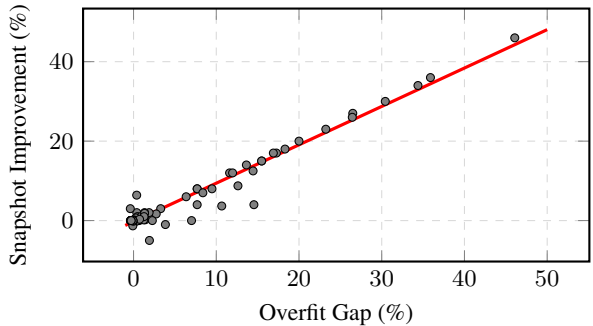

Figure ~\ref{fig:overfit-scatter} shows the correlation between the performance improvement of our our snapshot ensemble and the overfit gap.
Across all 111 tasks, the snapshot ensemble improvement increases monotonically with the over-fit gap (Spearman correlation $\rho = 0.97$), showing that the snapshot ensemble is most valuable when the baseline overfits.

While this analysis highlights the snapshot ensemble's value in reducing overfitting, it implicitly assumes the ensemble reliably aggregates multiple hypotheses of varying quality. 
To test this assumption, we next perform a sensitivity analysis.
We analyse two scenarios in detail.
We analyse the difference between the ensemble and the single best-performing test-optimal hypothesis to see whether the ensemble misses out by not selecting the very best individual hypothesis.
We also analyse the difference between the ensemble and the single hypothesis with the poorest test-set performance to see whether the ensemble can reliably avoid catastrophic decisions caused by bad hypotheses.

Figure \ref{fig:delta-ensemble-subplots} shows that the test-optimal-ensemble curve remains extremely close to the zero line for all cost functions. 
This result indicates that the ensemble consistently achieves nearly the same generalisation performance as the best individual snapshot identified at test time, regardless of the degree of overfitting. Secondly, the worst-snapshot-ensemble curve is substantially negative across almost every task. This shows that even when some snapshots generalise poorly, the ensemble effectively mitigates their negative impact, maintaining robust overall performance. 

Taken together, these results demonstrate that snapshot ensembles reliably approximate the best possible individual snapshot performance while safeguarding against significant accuracy losses from poorly generalising hypotheses.

Overall, our empirical results suggest that the answer to \textbf{Q1} is that snapshot ensembles consistently improve generalisation.


\subsubsection{Q2. Can Finding Multiple Optimal Hypotheses Improve Snapshot Ensemble Generalisation?}
The task‐level differences are not normally distributed. 
Therefore, a Wilcoxon signed‐rank test confirms ($p<0.05$)  that adding multiple optimal hypotheses to our snapshot ensemble improves accuracy on 7/111 (6\%) of tasks.
The rest are unchanged.
The mean change across 111 tasks is $0.03\%$ and the median is $0\%$.
Most improvement ($5/7$) come from the Alzheimer datasets with the \textsc{mdl} cost function.
Overall, these results suggest that the answer to \textbf{Q2} is that adding multiple optimal hypotheses alone yields no substantial advantage over using the snapshot pool with a single optimal hypothesis.

\subsubsection{Q3. Can Including Non-optimal Hypotheses Improve Snapshot Ensemble Generalisation?}
A Wilcoxon signed-rank test on the task-level accuracy differences (which are non-normally distributed) shows no significant difference when learning with and without non-optimal hypotheses.
Although the results are statistically insignificant, we note for completeness that the mean accuracy change is $0.08\%$ and the median is $0\%$.
Overall, these results suggest that the answer to \textbf{Q3} is that including non-optimal hypotheses does not yield a statistically significant accuracy improvement over including only optimal hypotheses.

\subsubsection{Q4. What is the Overhead of Snapshot Ensembles?}




Table~\ref{tab:overhead-ratio} shows the overhead of our snapshot ensemble approach.
The results show that it is less than 1\% overhead among all cost functions.
These results suggest that the answer \textbf {Q4} is that the overhead of our approach is effectively zero.

\begin{table}
\centering
\caption{Mean computational overhead of our symbolic snapshot ensemble relative to a baseline.
.}
\label{tab:overhead-ratio}
\begin{tabular}{@{}ll@{}}
\text{Cost function} & \text{Overhead ratio} \\
\midrule
\textsc{errorsize}  & 0.6 ± 0.7 \\
\textsc{lexfnsize}   & 0.1 ± 0.1 \\
\textsc{mdl}         & 0.2 ± 0.2 \\

\end{tabular}
\end{table}

\subsubsection{Q5. How does Our Snapshot Ensemble Method Compare to the Existing ILP Bagging Method? }

\begin{figure}[!b]
\centering
\begin{tikzpicture}

\begin{groupplot}[
  group style={
      group size=1 by 3,
      vertical sep=1cm,
      x descriptions at=edge bottom,
      y descriptions at=edge left
  },
  width=\columnwidth,
  height=.50\columnwidth,
  xmin=-1, xmax=13,     
  ymin=-5, ymax=10,
  xtick=\empty,
  xlabel={Task},
  every axis plot/.append style={mark size=0.8pt},
]

\nextgroupplot[
      title={Cost Function: \textsc{MDL}},
      ylabel={Accuracy difference (\%)},
            xmin=-1, xmax=13,   
      ymin=-5, ymax=8
]
    \addplot [only marks, mark=*, black, fill opacity=0.6,
            error bars/.cd, y dir=both, y explicit,
            error bar style={color=red}]
      table [x=n, y=mean, y error minus=minus, y error plus=plus]
      {diff_bagging_ranked_mdl.txt};
  \addplot [dashed, domain=0:24, samples=2, color=black] {0};

\nextgroupplot[
      title={Cost Function: \textsc{Errorsize}},
      ylabel={}
]
  \addplot [only marks, mark=*, black, fill opacity=0.6,
            error bars/.cd, y dir=both, y explicit,
            error bar style={color=red}]
      table [x=n, y=mean, y error minus=minus, y error plus=plus]
      {diff_bagging_ranked_errorsize.txt};
\addplot [dashed, domain=0:24, samples=2, color=black] {0};

\nextgroupplot[
      title={Cost Function: \textsc{Lexfnsize}},
      ylabel={},
      xticklabel style={font=\scriptsize},
      xmin=-1, xmax=13,   
      ymin=-10, ymax=49
]
  \addplot [only marks, mark=*, black, fill opacity=0.6,
            error bars/.cd, y dir=both, y explicit,
            error bar style={color=red}]
      table [x=n, y=mean, y error minus=minus, y error plus=plus]
      {diff_bagging_ranked_lex.txt};
\addplot [dashed, domain=0:24, samples=2, color=black] {0};

\end{groupplot}
\end{tikzpicture}
\caption{Task-level difference in accuracy (our symbolic ensemble's predictive accuracy minus the bagging's) across datasets. 
}
\label{fig:q1_single_column_three2}
\end{figure}
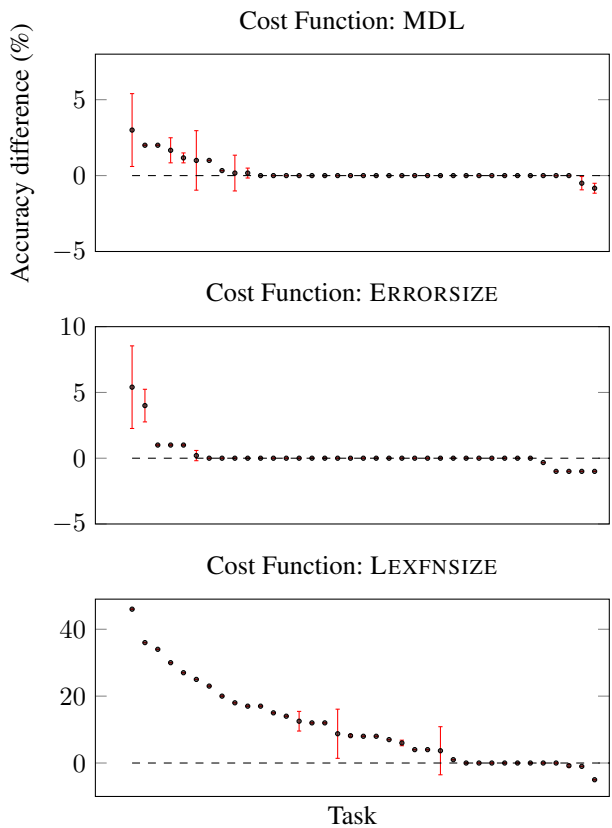

Figure \ref{fig:q1_single_column_three2} shows the differences in predictive accuracies when using snapshot ensembles compared to bagging.
Our snapshot ensemble matches or performs better than bagging on most tasks.  
A t-test confirms $(p < 0.05)$ that our approach improves accuracy on 42/111 (38\%) tasks and reduces accuracy on 10/111 (9\%) tasks, averaged over all cost functions.
For \textsc{ErrorSize} the mean improvement is $0.2\%\pm0.4\%$ and the ensemble underperforms on 5 tasks from Alzheimer, Wn18rr and drugdrug datasets.
For \textsc{mdl} the mean improvement is $0.3\%\pm0.3\%$, and underperforms on 2 tasks from the Alzheimer dataset.
For \textsc{LexFnSize} the mean improvement is $11.8\%\pm4.0\%$ and the ensemble loses on 3 tasks from Alzheimer and 1D-ARC datasets.

Table \ref{tab:bagging‐vs‐snapshot‐ci} shows the mean performance across all datasets.
The table shows that our snapshot method outperforms bagging for all three cost functions.
The advantage is most pronounced for \textsc{lexfnsize}, with a 12\% improvement.
For \textsc{mdl} the improvement is 1\%.
For \textsc{errorsize} both methods perform identically once values are rounded to whole percentages.

In terms of computational overhead, running three bootstrap replicates for bagging extends the total time to about 180 minutes.
In other words, bagging is 200\% slower.
Additionally, the running cost of bagging grows linearly with the number of bootstrap samples.
By contrast, our snapshot ensemble adds less than 1\% overhead.

Overall, these results suggest that the answer to \textbf{Q5} is that our snapshot ensemble matches or exceeds the predictive performance of bagging while requiring substantially less computational cost.

\begin{table}[h]
\centering
\caption{Mean predictive accuracy across all tasks.  Bold indicates the best accuracy for each cost function.
}
\label{tab:bagging‐vs‐snapshot‐ci}
\begin{tabular}{@{}l|ll@{}}
Cost function & Bagging & Snapshot \\
\midrule
\textsc{errorsize} & 78\,$\pm$\,5 & \textbf{78\,$\pm$\,5} \\
\textsc{lexfnsize}  & 59\,$\pm$\,6 & \textbf{71\,$\pm$\,5} \\
\textsc{mdl}        & 84\,$\pm$\,4 & \textbf{85\,$\pm$\,4} \\
\end{tabular}
\end{table}





\section{Conclusions and Limitations}
We introduced snapshot ensembles for ILP. 
In this approach, we train an ILP algorithm once and save the intermediate hypotheses it generates during training. 
We then combine these hypotheses using a MDL weighting scheme. 
Our experiments on 111 diverse tasks show that symbolic snapshot ensembles consistently improve generalisation with virtually no additional computational cost. 
The approach has the greatest benefit in scenarios where overfitting is most severe. 
The results show that our snapshot ensemble approach matches or surpasses the predictive performance of standard bagging methods, while requiring significantly less computational cost.
Overall, this work shows that a simple idea can significantly boost generalisation without sacrificing efficiency.

\subsection*{Limitations}

\paragraph{ILP systems.}
We have shown that our snapshot ensemble is beneficial for an off-the-shelf ILP system. 
Our snapshot ensemble is agnostic to the choice of ILP learner because it only needs access to intermediate hypotheses.
Assuming an optimal learning system, the same cost function, and the same hypothesis space, the cost of a training optimal hypothesis with a different system should be the same.
However, different systems may find different intermediate hypotheses due to variations in their search strategies. 
Consequently, the diversity and quality of snapshots may vary across systems.
Showing the benefits of our approach with other ILP systems is a direction for future work.

\paragraph{Hyperparameters.}
For efficiency, we keep the weighting hyperparameters $\alpha$ and $\beta$ fixed instead of resetting them for every dataset or task. 
The $\alpha:\beta$ ratios balanced empirical error against the MDL penalty.
We selected the ratio with the highest average validation accuracy across all tasks. 
A single global setting, however, may be sub‑optimal for individual datasets. 
Future work should investigate adaptive methods or task‑specific priors that adjust $\alpha$ and $\beta$ automatically.

\paragraph{Hypothesis selection and weighting strategy.}
We purposely use a simple approach to demonstrate the idea of snapshot ensembles. 
Our work opens two new lines of research. 
One is exploring better methods to find and select hypotheses to add to the snapshot pool.
The other is to find better ways to combine them in an ensemble.


\bibliography{neurips_bib}

\begin{thebibliography}{36}
\providecommand{\natexlab}[1]{#1}

\bibitem[{Blockeel and {De Raedt}(1998)}]{tilde}
Blockeel, H.; and {De Raedt}, L. 1998.
\newblock Top-Down Induction of First-Order Logical Decision Trees.
\newblock \emph{Artif. Intell.}, 101(1-2): 285--297.

\bibitem[{Chen, Lundberg, and Lee(2017)}]{chen2017checkpoint}
Chen, H.; Lundberg, S.; and Lee, S.-I. 2017.
\newblock Checkpoint ensembles: Ensemble methods from a single training process.
\newblock \emph{arXiv preprint arXiv:1710.03282}.

\bibitem[{Chollet(2019)}]{arc}
Chollet, F. 2019.
\newblock On the Measure of Intelligence.
\newblock \emph{CoRR}, abs/1911.01547.

\bibitem[{Cropper and Dumancic(2022)}]{ilpintro}
Cropper, A.; and Dumancic, S. 2022.
\newblock Inductive Logic Programming At 30: {A} New Introduction.
\newblock \emph{J. Artif. Intell. Res.}, 74: 765--850.

\bibitem[{Cropper, Evans, and Law(2020)}]{iggp}
Cropper, A.; Evans, R.; and Law, M. 2020.
\newblock Inductive general game playing.
\newblock \emph{Mach. Learn.}, 109(7): 1393–1434.

\bibitem[{Cropper and Morel(2021)}]{popper}
Cropper, A.; and Morel, R. 2021.
\newblock Learning programs by learning from failures.
\newblock \emph{Mach. Learn.}, 110(4): 801--856.

\bibitem[{Davis et~al.(2005)Davis, Burnside, de~Castro~Dutra, Page, and Costa}]{davis2005integrated}
Davis, J.; Burnside, E.; de~Castro~Dutra, I.; Page, D.; and Costa, V.~S. 2005.
\newblock An integrated approach to learning Bayesian networks of rules.
\newblock In \emph{European Conference on Machine Learning}, 84--95. Springer.

\bibitem[{de~Castro~Dutra et~al.(2003)de~Castro~Dutra, Page, Santos~Costa, and Shavlik}]{ilpbagging2003}
de~Castro~Dutra, I.; Page, D.; Santos~Costa, V.; and Shavlik, J. 2003.
\newblock An empirical evaluation of bagging in inductive logic programming.
\newblock In \emph{Inductive Logic Programming: 12th International Conference, ILP 2002 Sydney, Australia, July 9--11, 2002 Revised Papers 12}, 48--65. Springer.

\bibitem[{{De Raedt}(2008)}]{luc:book}
{De Raedt}, L. 2008.
\newblock \emph{Logical and relational learning}.
\newblock ISBN 978-3-540-20040-6.

\bibitem[{{De Raedt} et~al.(2015){De Raedt}, Dries, Thon, den Broeck, and Verbeke}]{probfoil}
{De Raedt}, L.; Dries, A.; Thon, I.; den Broeck, G.~V.; and Verbeke, M. 2015.
\newblock Inducing Probabilistic Relational Rules from Probabilistic Examples.
\newblock In \emph{Proceedings of the Twenty-Fourth International Joint Conference on Artificial Intelligence, {IJCAI} 2015, Buenos Aires, Argentina, July 25-31, 2015}, 1835--1843.

\bibitem[{Dettmers et~al.(2018)Dettmers, Minervini, Stenetorp, and Riedel}]{dettmers2018convolutional}
Dettmers, T.; Minervini, P.; Stenetorp, P.; and Riedel, S. 2018.
\newblock Convolutional 2d knowledge graph embeddings.
\newblock In \emph{Proceedings of the AAAI conference on artificial intelligence}, volume~32.

\bibitem[{Dhami et~al.(2018)Dhami, Kunapuli, Das, Page, and Natarajan}]{dhami2018drug}
Dhami, D.~S.; Kunapuli, G.; Das, M.; Page, D.; and Natarajan, S. 2018.
\newblock Drug-drug interaction discovery: kernel learning from heterogeneous similarities.
\newblock \emph{Smart Health}, 9: 88--100.

\bibitem[{Dietterich(2000)}]{dietterich2000ensemble}
Dietterich, T.~G. 2000.
\newblock Ensemble methods in machine learning.
\newblock In \emph{International workshop on multiple classifier systems}, 1--15. Springer.

\bibitem[{Evans and Grefenstette(2018)}]{dilp}
Evans, R.; and Grefenstette, E. 2018.
\newblock Learning Explanatory Rules from Noisy Data.
\newblock \emph{J. Artif. Intell. Res.}, 61: 1--64.

\bibitem[{Hoche and Wrobel(2001)}]{hoche2001relational}
Hoche, S.; and Wrobel, S. 2001.
\newblock Relational learning using constrained confidence-rated boosting.
\newblock In \emph{International Conference on Inductive Logic Programming}, 51--64. Springer.

\bibitem[{Hocquette et~al.(2024)Hocquette, Niskanen, J{\"{a}}rvisalo, and Cropper}]{maxsynth}
Hocquette, C.; Niskanen, A.; J{\"{a}}rvisalo, M.; and Cropper, A. 2024.
\newblock Learning {MDL} Logic Programs from Noisy Data.
\newblock In Wooldridge, M.~J.; Dy, J.~G.; and Natarajan, S., eds., \emph{Thirty-Eighth {AAAI} Conference on Artificial Intelligence, {AAAI} 2024, Thirty-Sixth Conference on Innovative Applications of Artificial Intelligence, {IAAI} 2024, Fourteenth Symposium on Educational Advances in Artificial Intelligence, {EAAI} 2014, February 20-27, 2024, Vancouver, Canada}, 10553--10561. {AAAI} Press.

\bibitem[{Huang et~al.(2017)Huang, Li, Pleiss, Liu, Hopcroft, and Weinberger}]{huang2017snapshot}
Huang, G.; Li, Y.; Pleiss, G.; Liu, Z.; Hopcroft, J.~E.; and Weinberger, K.~Q. 2017.
\newblock Snapshot ensembles: Train 1, get m for free.
\newblock \emph{arXiv preprint arXiv:1704.00109}.

\bibitem[{Jiang and Colton(2006)}]{jiang2006boosting}
Jiang, N.; and Colton, S. 2006.
\newblock Boosting descriptive ILP for predictive learning in bioinformatics.
\newblock In \emph{International Conference on Inductive Logic Programming}, 275--289. Springer.

\bibitem[{King, Sternberg, and Srinivasan(1995)}]{alzheimer}
King, R.~D.; Sternberg, M.~J.; and Srinivasan, A. 1995.
\newblock Relating chemical activity to structure: an examination of ILP successes.
\newblock \emph{New Generation Computing}, 13: 411--433.

\bibitem[{Kramer(2001)}]{kramer2001demand}
Kramer, S. 2001.
\newblock Demand-driven construction of structural features in ILP.
\newblock In \emph{International Conference on Inductive Logic Programming}, 132--141. Springer.

\bibitem[{Law, Russo, and Broda(2014)}]{ilasp}
Law, M.; Russo, A.; and Broda, K. 2014.
\newblock Inductive Learning of Answer Set Programs.
\newblock In Ferm{\'{e}}, E.; and Leite, J., eds., \emph{Logics in Artificial Intelligence - 14th European Conference, {JELIA} 2014, Funchal, Madeira, Portugal, September 24-26, 2014. Proceedings}, volume 8761 of \emph{Lecture Notes in Computer Science}, 311--325. Springer.

\bibitem[{Liu et~al.(2021)Liu, Chen, Atashgahi, Chen, Sokar, Mocanu, Pechenizkiy, Wang, and Mocanu}]{liu2021deep}
Liu, S.; Chen, T.; Atashgahi, Z.; Chen, X.; Sokar, G.; Mocanu, E.; Pechenizkiy, M.; Wang, Z.; and Mocanu, D.~C. 2021.
\newblock Deep ensembling with no overhead for either training or testing: The all-round blessings of dynamic sparsity.
\newblock \emph{arXiv preprint arXiv:2106.14568}.

\bibitem[{Lloyd(2012)}]{lloyd:book}
Lloyd, J.~W. 2012.
\newblock \emph{Foundations of logic programming}.
\newblock Springer Science \& Business Media.

\bibitem[{Muggleton(1991)}]{mugg:ilp}
Muggleton, S. 1991.
\newblock Inductive Logic Programming.
\newblock \emph{New Generation Computing}, 8(4): 295--318.

\bibitem[{Muggleton(1995)}]{progol}
Muggleton, S. 1995.
\newblock Inverse Entailment and Progol.
\newblock \emph{New Generation Comput.}, 13(3{\&}4): 245--286.

\bibitem[{Muggleton, Lin, and Tamaddoni{-}Nezhad(2015)}]{metagold}
Muggleton, S.~H.; Lin, D.; and Tamaddoni{-}Nezhad, A. 2015.
\newblock Meta-interpretive learning of higher-order dyadic datalog: predicate invention revisited.
\newblock \emph{Mach. Learn.}, 100(1): 49--73.

\bibitem[{Quinlan(1990)}]{foil}
Quinlan, J.~R. 1990.
\newblock Learning Logical Definitions from Relations.
\newblock \emph{Mach. Learn.}, 5: 239--266.

\bibitem[{Quinlan(1996)}]{quinlan1996boosting}
Quinlan, J.~R. 1996.
\newblock Boosting first-order learning.
\newblock In \emph{International workshop on algorithmic learning theory}, 143--155. Springer.

\bibitem[{Rissanen(1978)}]{mdl}
Rissanen, J. 1978.
\newblock Modeling by shortest data description.
\newblock \emph{Autom.}, 14(5): 465--471.

\bibitem[{Salvini, Aguilar, and Dutra(2007)}]{salvini2007skimmed}
Salvini, R.; Aguilar, E.; and Dutra. 2007.
\newblock Skimmed classifiers.
\newblock In \emph{Work-in-Progress Proceedings of the 2007 International Conference on Inductive Logic Programming,(Oregon, USA)}. Citeseer.

\bibitem[{Srinivasan(2001)}]{aleph}
Srinivasan, A. 2001.
\newblock The {ALEPH} manual.
\newblock \emph{Machine Learning at the Computing Laboratory, Oxford University}.

\bibitem[{Wasay et~al.(2020)Wasay, Hentschel, Liao, Chen, and Idreos}]{wasay2020mothernets}
Wasay, A.; Hentschel, B.; Liao, Y.; Chen, S.; and Idreos, S. 2020.
\newblock Mothernets: Rapid deep ensemble learning.
\newblock \emph{Proceedings of Machine Learning and Systems}, 2: 199--215.

\bibitem[{Whitaker and Whitley(2022)}]{whitaker2022prune}
Whitaker, T.; and Whitley, D. 2022.
\newblock Prune and tune ensembles: low-cost ensemble learning with sparse independent subnetworks.
\newblock In \emph{Proceedings of the AAAI Conference on Artificial Intelligence}, volume~36, 8638--8646.

\bibitem[{Xie, Xu, and Chuang(2013)}]{xie2013horizontal}
Xie, J.; Xu, B.; and Chuang, Z. 2013.
\newblock Horizontal and vertical ensemble with deep representation for classification.
\newblock \emph{arXiv preprint arXiv:1306.2759}.

\bibitem[{Xu et~al.(2023)Xu, Li, Vaezipoor, Sanner, and Khalil}]{onedarc}
Xu, Y.; Li, W.; Vaezipoor, P.; Sanner, S.; and Khalil, E.~B. 2023.
\newblock {LLM}s and the Abstraction and Reasoning Corpus: Successes, Failures, and the Importance of Object-based Representations.
\newblock \emph{CoRR}, abs/2305.18354.

\bibitem[{Yan et~al.(2024)Yan, Natarajan, Joshi, Khardon, and Tadepalli}]{yan2024explainable}
Yan, S.; Natarajan, S.; Joshi, S.; Khardon, R.; and Tadepalli, P. 2024.
\newblock Explainable models via compression of tree ensembles.
\newblock \emph{Machine Learning}, 113(3): 1303--1328.

\end{thebibliography}
\end{document}